# Evaluating Biomedical Reranking for LLM-Based Question Answering over Longitudinal Clinical Notes

*Maryam Shahbaz Ali[1], Laura B. Strachan[1,2], Caitlin Sherman[1], Mark Kovler[1], Eleanor Mackey[1,3], Syed Muhammad Anwar[1,3]*

[1] Children's National Hospital, Washington, DC, USA
[2] University of Florida, Gainesville, FL, USA
[3] George Washington University, Washington DC, USA

## ABSTRACT

Patient-specific clinical question answering requires locating the right evidence within long, heterogeneous longitudinal clinical records in which relevant facts may be scattered across encounters, repeated in copied-forward notes, or expressed using different clinical terminology. We evaluated whether biomedical reranking can improve evidence selection and downstream answer quality in a locally deployed retrieval-augmented generation pipeline for longitudinal clinical notes. The pipeline combines PubMedBERT dense retrieval, BM25 lexical retrieval, weighted reciprocal-rank fusion, and MedCPT cross-encoder reranking. Across 1,000 open- and closed-ended question-answer pairs from a cohort of 200 bariatric surgery patients, reranking increased exact source-chunk retrieval within the top 10 items, Hit@10 from 46.6% to 60.6% and mean reciprocal rank from 0.2371 to 0.3252. With Qwen3-8B generation, local judge-assessed answer correctness increased from 44.8% to 48.6%. These results show that biomedical reranking can improve the placement of relevant clinical evidence within a limited context window, although gains in retrieval do not translate proportionally into gains in answer correctness.

## 1. INTRODUCTION

Patient-specific clinical question answering (QA) differs from general medical QA because the required evidence is often buried within a single patient's longitudinal record rather than contained in model pretraining. Relevant facts may be distributed across encounters, repeated in copied-forward notes, or expressed using different clinical terminology. This creates a retrieval problem in which the system must identify the most informative evidence from a large, redundant record while operating within a limited context window.

Retrieval-augmented generation (RAG) provides a mechanism for supplying patient-specific evidence to a language model, but the usefulness of the final answer depends heavily on which passages are retrieved and how they are ranked. Towards this goal, MIRAGE benchmark and MedRAG toolkit were introduced to systematically evaluate retrieval configurations for the medical QA [1]. In longitudinal clinical records, this distinction is particularly important because retrieving the relevant evidence and correctly interpreting that evidence are separate stages of the pipeline.

Dense retrieval can capture semantically related text, whereas lexical retrieval rewards direct term overlap. These complementary signals can be combined using reciprocal-rank fusion (RRF) [2], after which a cross encoder can jointly score the question and candidate passage. MedCPT provides a biomedical retrieval and reranking framework trained from PubMed search interactions [3]. Recent work has extended retrieval augmentation to electronic health record applications beyond general medical QA. RAG was applied to EHR summarization and clinical information extraction [4], while RAM-EHR was introduced to augment clinical prediction with retrieved medical knowledge [5]. Recent studies have further examined RAG for biomedical applications, longitudinal EHR reasoning, factual-aware medical QA, and evidence-grounded patient QA [8]-[12]. These studies demonstrate the broader value of retrieval augmentation, but do not directly evaluate whether biomedical cross-encoder reranking improves evidence of selection for patient-specific question answering over longitudinal clinical notes.

We evaluate this approach using a frozen set of 1,000 QA pairs generated from 200 bariatric surgery patients, with retrieval restricted to each patient's longitudinal record and clinical inference performed locally. This work makes three contributions: 1) evaluates biomedical cross-encoder reranking for longitudinal clinical QA, 2) quantifies the relationship between improved evidence retrieval and downstream answer correctness, including separate analysis of open- and closed-ended QA, and 3) demonstrates a fully local, patient-restricted RAG evaluation incorporating retrieval, generation, and identifier screening.

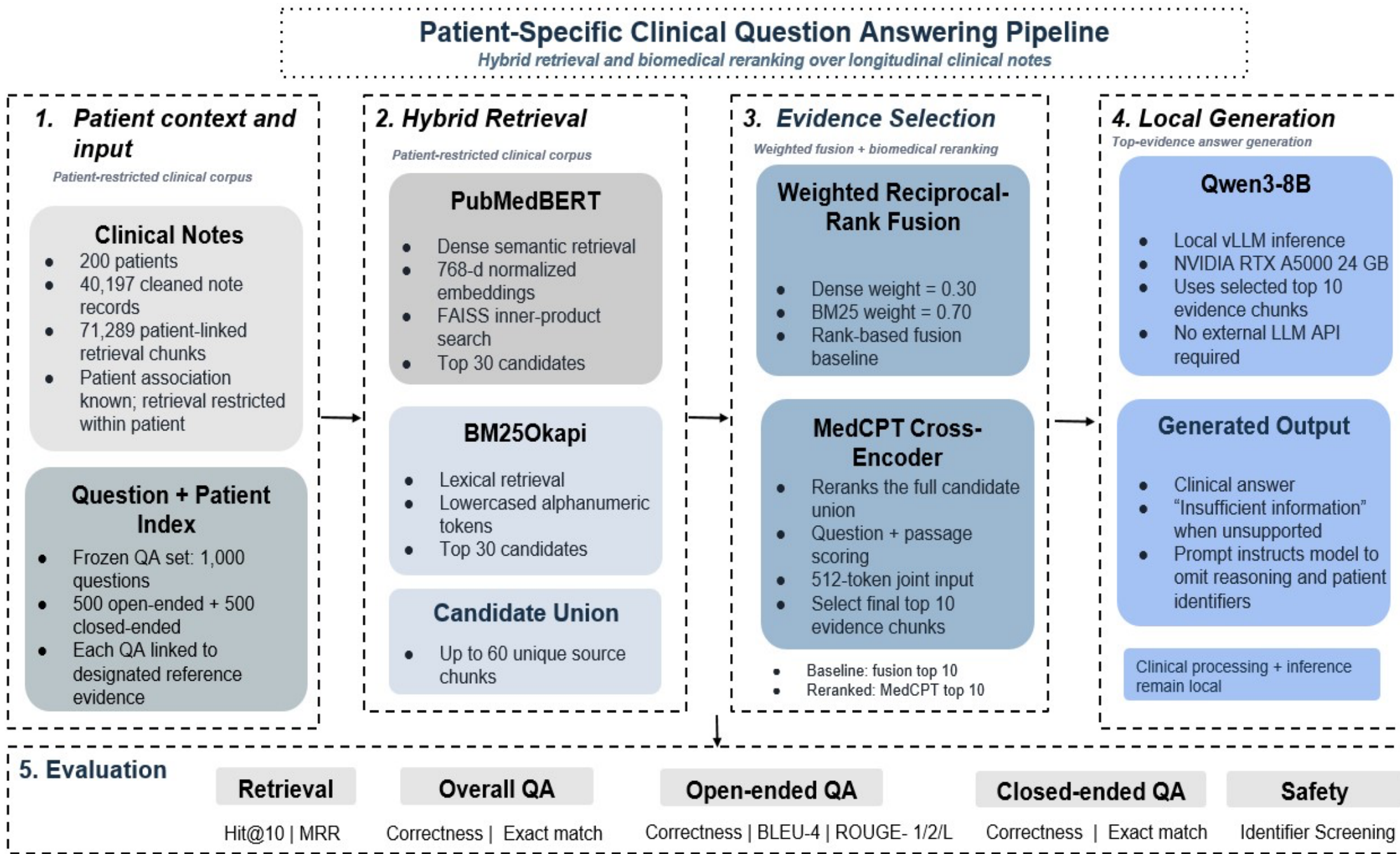

**Fig. 1**. Patient-specific clinical question answering using hybrid PubMedBERT and BM25 retrieval, weighted rank fusion, MedCPT reranking, and local Qwen 3-8B generation. Retrieval and answer quality are evaluated separately.

## 2. METHODS

An overview of the proposed method is presented in Figure 1, with details about each component in the following text.

### 2.1. Dense and lexical candidate generation

Questions and retrieval chunks were encoded using NeuML's PubMedBERT base embeddings checkpoint through Sentence Transformers [6]. The 768-dimensional vectors are normalized to unit length. Dense retrieval used a FAISS inner-product index within each patient's corpus, yielding cosine-similarity rankings. Cached corpus embeddings were verified against the corpus; both contained 71,289 rows with matching source-identifier ordering.

Lexical retrieval used the BM25Okapi implementation from the rank_bm25 Python package, based on the Okapi BM25 ranking function [13]. Text was lowercased and tokenized into alphanumeric sequences using the regular expression [a-z 0-9] +. Each retriever returned up to 30 candidates, producing a union of at most 60 unique source chunks.

For a candidate chunk c and query q, weighted reciprocal rank fusion is assigned as,

$$s_{fusion}(q,c) = \frac{0.301[c \in D_q]}{60 + r_D(c)} + \frac{0.701[c \in B_q]}{60 + r_B(c)}, \qquad (1)$$

where Dq and Bq are the truncated dense and BM25 rankings, respectively, and ranks begin at one. The plus sign denotes the sum of the dense and BM25 reciprocal-rank contributions. The 0.30/0.70 weights were selected during development based on retrieval performance, with BM25 weighted more heavily because it performed better on this cohort. The denominator constant 60 is the standard RRF offset. Candidates absent from a ranking receive zero contribution from that retriever. The coefficients are rank-fusion weights rather than weights on raw PubMedBERT or BM25 scores.

### 2.2. Biomedical reranking and generation

The reranked condition uses the ncbi/MedCPT cross-encoder [3] checkpoint to score the full union of up to 30 dense and 30 BM25 candidates for each question. Question–passage pairs were processed in batches of eight, with the question followed by the candidate passage. The combined input is limited to 512 tokens, with truncation applied only to the passage. Candidates are ordered by descending model score, and the ten highest-ranked chunks are selected. Tied scores retain the weighted reciprocal-rank fusion order. The reranker runs locally on the CPU using cached model files. This comparison uses the original cached PubMedBERT embeddings rather than the token-window variant. Because all candidates are reranked, the fusion weights affect the final ranking only when reranker scores are tied.

Qwen3-8B [7] generates answers with either no retrieved context or the selected top ten chunks. Previously generated no-RAG answers were reused for the reranked experiment, while RAG answers were regenerated. The clinical generation was served through a local vLLM endpoint on an NVIDIA RTX A5000 with 24GB memory; reranking ran on CPU. Public reranker weights were downloaded before local clinical processing. The local execution is a deployment property, not a guaranteed text containing no identifiers. The generation prompt instructs the model to answer only from the information provided and to return "Insufficient information", when an answer is unsupported. Closed-ended answers are normally limited to 20 words, while open-ended answers are requested in one to three sentences. The prompt also instructs the model to omit reasoning and patient identifiers. Retrieved passages were numbered in rank order and separated by blank lines.

## 3. EXPERIMENTS AND RESULT

### 3.1. Dataset

The retrieval corpus was constructed from longitudinal clinical note records belonging to the 200-patient evaluation cohort who underwent bariatric surgery at our clinical center. Data are collected under an approved IRB study. Clinical note text was cleaned by normalizing whitespace, removing records with empty text or invalid timestamps, and eliminating exact duplicates, yielding 40,197 cleaned note records. Each cleaned note was converted into one or more retrieval chunks, defined as contiguous segments of clinical text that serve as the searchable units of the retrieval system. Chunking was character-based with a target length of approximately 1,600 characters. Notes containing fewer than 250 characters after cleaning were excluded as very short text fragments, while notes containing 250-2,200 characters were retained as single chunks. Longer notes were split at newline boundaries, while trailing text carried into subsequent chunks to preserve local context. Duplicate chunks were removed within each patient after lowercasing and whitespace normalization. After removal of 7,925 duplicate chunks, the final retrieval corpus contained 71,289 patient-linked retrieval chunks.

Retrieval was patient restricted: each question searched only chunks belonging to the corresponding patient. Each chunk retained a source identifier linking it to the designated reference evidence used for evaluation; source identifiers were not used as retrieval features. This setting assumes that patient identity is already known and does not evaluate cross-patient record discovery.

### 3.2. Evaluation questions and reference evidence

Patients were eligible if they had at least 20 non-empty note records with valid timestamps and at least five distinct encounters. The frozen cohort matched a random sample of 200 eligible patients selected with seed 42. The frozen evaluation set contains five questions per patient: 1,000 total, comprising 500 open-ended and 500 closed-ended questions. Each row contains a patient index, question, reference answer, and a designated source chunk.

An initial automated quality check using Qwen3-8B [7] classified 993 questions as pass, four requiring reviews, and three as fail. Subsequent checks covered 117 repaired rows, and six corrected rows, all of which passed their rechecks. One initially review-flagged question was absent from both later checks; hence its resolution could not be established from the retained quality-check records. These automated checks do not establish independent human validation. Several configurations were examined on the same 200 patient cohort. These findings therefore describe performance during development and not just the independently held-out generalizations. No independent held-out test split or manual validation is claimed in this study.

### 3.3. Retrieval and answer metrics

The exact source Hit@10 measures whether the designated source chunk occurs among the ten retrieved chunks. The reciprocal rank is 1/r at the first occurrence of that chunk and zero when absent from the searched candidate list. Reported mean reciprocal rank (MRR) is therefore candidate-limited: the individual retrievers search up to 30 items, while the fused and reranked lists can contain up to 60. An exact source miss therefore does not prove that all the retrieved passages are irrelevant; overlapping or repeated notes may contain equivalent evidence.

Automated semantic correctness is the primary answer metric against the frozen reference answer. A local Qwen3-8B [7] judge evaluates blinded candidate A/B answers, with candidate assignment determined by a SHA256 hash of the QA identifier. The judge receives the question, question type, reference answer, and both candidate answers. It is instructed to assess each candidate independently, allowing both to be correct and incorrect. Closed-ended answers must provide the requested fact, allowing equivalent wording, standard abbreviations, and equivalent numerical values or units. Open-ended answers must contain the essential information without material errors or contradictions. Insufficient information is scored as incorrect when the reference provides an answer. Retrieved passages are not supplied to the judge, so this assessment measures agreement with the reference rather than grounding in the retrieved context. Answer performance was additionally evaluated separately for the 500 open-ended and 500 closed-ended questions. For the open-ended subset, BLEU-4 and ROUGE-1, ROUGE-2, and ROUGE-L F1 scores were computed as complementary text-overlap metrics. Normalized exact match is a secondary metric. Candidate and reference answers are lowercased, characters other than letters, digits, periods, percent signs, plus signs, hyphens, and slashes are replaced with spaces, and whitespace is collapsed before comparison.

**Table 1.** Exact source retrieval on 1,000 questions. MRR uses each method's searched candidate list.

| Method | Hit@10 (%) | MRR |
|---|---|---|
| PubMedBERT dense | 37.9 | 0.1621 |
| BM25 | 43.7 | 0.2244 |
| Weighted hybrid | 46.6 | 0.2371 |
| Hybrid with MedCPT | 60.6 | 0.3252 |

We estimated 95% confidence intervals using 5,000 patient-cluster bootstrap replicates with random seed 42. Patients were sampled with replacement, retaining their associated questions. For the RAG versus no-RAG comparison, each replicate measured RAG accuracy minus no-RAG accuracy. For the weighted hybrid versus MedCPT comparison, results were paired by question identifier, and patient assignments and question types were checked for consistency. Differences in Hit@10 and automated answer correctness were calculated as MedCPT minus weighted hybrid. Interval bounds were the 2.5th and 97.5th percentiles of the bootstrap differences. Patient-cluster intervals are the principal uncertainty summaries. Row-level exact paired binomial tests were also calculated, but they do not account for within-patient dependence. Comparisons between RAG variants use labels from separate automated judging runs and are exploratory.

An automated local audit screened generated answers for possible patient names and other identifiers. We report flagged counts rather than confirmed disclosures; all flags require manual review.

**Table 2.** Automated answer correctness and exact match (%). No-RAG answers were reused but judged separately in each comparison.

| Condition | Correctness (%) | Exact match (%) |
|---|---|---|
| No-RAG, baseline judge run | 6.4 | 0.0 |
| Weighted-hybrid RAG | 44.8 | 6.2 |
| No-RAG, reranked judge run | 6.3 | 0.0 |
| MedCPT-reranked RAG | 48.6 | 8.0 |
| Open-ended, weighted hybrid RAG | 49.6 | 2.4 |
| Open-ended, MedCPT-reranked RAG | 51.8 | 1.8 |
| Closed-ended, weighted-hybrid RAG | 40.0 | 10.0 |
| Closed-ended, MedCPT-reranked RAG | 45.4 | 14.2 |

### 3.4. Evidence retrieval

Table 1 summarizes retrieval performance. Weighted fusion achieved 46.6% Hit@10, compared with 37.9% for dense retrieval and 43.7% for BM25. MedCPT reranking increased Hit@10 to 60.6% and MRR to 0.3252. Relative to weighted fusion, 171 questions changed from miss to hit and 31 from hit to miss. The net increase was 140 questions, or 14.0 percentage points (95% patient-cluster CI, 11.4–16.6).

The union of the top 30 dense and top 30 BM25 candidates contained the designated chunk for 778 questions (77.8%). This is a candidate coverage bound for exact-source retrieval under this pool, not an answer-accuracy bound. Reranking placed 606 of these targets in the top ten; 172 remained below rank ten, while 222 were absent from the candidate pool.

### 3.5. Downstream question answering

MedCPT reranking increased automated answer correctness from 44.8% with weighted-hybrid RAG to 48.6%, an absolute gain of 3.8 percentage points (95% patient-cluster CI, 1.1-6.5). Across the two judging runs, 123 questions changed from incorrect to correct and 85 changed in the opposite direction.

When stratified by question type, correctness increased from 49.6% to 51.8% for open-ended questions and from 40.0% to 45.4% for closed-ended questions. For open-ended QA, BLEU-4 increased from 0.1813 to 0.1883, ROUGE-1 F1 from 0.4300 to 0.4545, ROUGE-2 F1 from 0.2702 to 0.2863, and ROUGE-L F1 from 0.3910 to 0.4093. In the reranked condition, correctness was 63.4% for questions in which the designated source was retrieved within the top ten and 25.9% for retrieval misses. However, 222 of 606 retrieval-hit questions were still judged incorrect, indicating that successful evidence retrieval did not guarantee a correct generated answer.

### 3.6. Window ablation and identifier screening

Tokenization identified 5,528 of the 71,289 chunks (7.75%) above the embedding model's 512-token limit. A 448-token overlapping-window ablation achieved 46.5% hybrid Hit @10 and 0.2372 MRR, providing no improvement over the original hybrid (46.6%, 0.2371); therefore, it was not used for reranking. The audit flagged any identifier in 44 weighted-hybrid RAG answers and 40 reranked RAG answers. No-RAG had 15 flagged answers in both experiments. Condition flags decreased from 56 to 48 questions. Among the 952 questions for which neither condition's answer was flagged by the identifier screen in the reranked evaluation, RAG correctness was 49.1%. Different flagged subsets preclude interpreting these filtered accuracies as a matched comparison between RAG variants. The flagged outputs require review before any release.

## 4. DISCUSSION

Biomedical reranking improved evidence selection by increasing Hit@10 by 14.0 percentage points, while automated answer correctness increased by 3.8 points overall. Stratified gains were 2.2 points for open-ended and 5.4 points for closed-ended QA. Remaining errors may reflect incomplete evidence, distracting context, generation errors, reference limitations, or judge errors.

Several limitations constrain interpretation. The same Qwen3-8B family generated and judged answers, the development cohort was reused for configuration selection, exact-source labels may undercount equivalent evidence, and patient-restricted retrieval assumes reliable patient association. Independent human evaluation and held-out validation are needed to establish generalization.

## 5. CONCLUSION

In a local patient-specific QA evaluation, MedCPT reranking improved exact source Hit@10 from 46.6% to 60.6% and automated answer correctness from 44.8% to 48.6%. The results support a distinction between evidence retrieval and downstream answer quality. Independent human evaluation and held-out validation are needed to establish generalization.

## REFERENCES

[1] G. Xiong, Q. Jin, Z. Lu, and A. Zhang, "Benchmarking retrieval-augmented generation for medicine," in Findings of the Association for Computational Linguistics: ACL 2024, 2024, pp. 6233–6251.

[2] G. V. Cormack, C. L. A. Clarke, and S. Büttcher, "Reciprocal rank fusion outperforms Condorcet and individual rank learning methods," in Proc. 32nd Int. ACM SIGIR Conf. Research and Development in Information Retrieval, 2009, pp. 758–759.

[3] Q. Jin et al., "MedCPT: Contrastive pre-trained transformers with large-scale PubMed search logs for zero-shot biomedical information retrieval," Bioinformatics, vol. 39, no. 11, Art. no. btad651, 2023.

[4] M. Alkhalaf, P. Yu, M. Yin, and C. Deng, "Applying generative AI with retrieval augmented generation to summarize and extract key clinical information from electronic health records," Journal of Biomedical Informatics, vol. 156, Art. no. 104662, 2024.

[5] R. Xu, W. Shi, Y. Yu, Y. Zhuang, B. Jin, M. D. Wang, J. C. Ho, and C. Yang, "RAM-EHR: Retrieval Augmentation Meets Clinical Predictions on Electronic Health Records," in Proc. 62nd Annual Meeting of the Association for Computational Linguistics (Volume 2: Short Papers), 2024, pp. 754–765.

[6] NeuML, "PubMedBERT Base Embeddings," Hugging Face model card, accessed Sep. 13, 2026.

[7] Qwen Team, "Qwen3 Technical Report," arXiv preprint arXiv:2505.09388, 2025.

[8] S. Liu, A. B. McCoy, and A. Wright, "Improving large language model applications in biomedicine with retrieval-augmented generation: a systematic review, meta-analysis, and clinical development guidelines," Journal of the American Medical Informatics Association, vol. 32, no. 4, pp. 605–615, 2025.

[9] S. Myers et al., "Evaluating retrieval-augmented generation versus long-context input for clinical reasoning over electronic health records," Journal of the American Medical Informatics Association, Art. no. ocag139, 2026.

[10] S. Liang, L. Zhang, H. Zhu, W. Wang, Y. He, and D. Zhou, "RGAR: Recurrence Generation-augmented Retrieval for Factual-aware Medical Question Answering," in Findings of the Association for Computational Linguistics: EMNLP 2025, 2025, pp. 4006–4033.

[11] A. Chouhan and M. Gertz, "heiDS at ArchEHR-QA 2025: From Fixed-k to Query-dependent-k for Retrieval Augmented Generation," in Proceedings of the 24th Workshop on Biomedical Language Processing (Shared Tasks), 2025, pp. 50–61.

[12] P. Kadusabe, A. Kaushik, and F. Lawless, "DKITNLP at ArchEHR-QA 2025: A Retrieval Augmented LLM Pipeline for Evidence-Based Patient Question Answering," in Proceedings of the 24th Workshop on Biomedical Language Processing (Shared Tasks), 2025, pp. 165–1.

[13] S. Robertson and H. Zaragoza, "The probabilistic relevance framework: BM25 and beyond," Foundations and Trends in Information Retrieval, vol. 3, no. 4, pp. 333–389, 2009.